\documentclass[a4paper,conference]{IEEEtran}
\usepackage{times}
\usepackage{epsfig}
\usepackage{graphicx}
\usepackage{amsmath}
\usepackage{amssymb}
\usepackage{pdfpages}
\usepackage{enumitem}
\usepackage{listings}

\usepackage{multirow}
\usepackage{adjustbox}

\ifCLASSINFOpdf
\else
\fi
\begin{document}
%
\title{ContCap: A Scalable Framework for \\
Continual Image Captioning}

\author{\IEEEauthorblockN{Giang Nguyen, Trung Tran, Tolcha Yalew, Daeyoung Kim}
\IEEEauthorblockA{School of Computing\\
KAIST\\
Daejeon, South Korea\\
\{dexter.nguyen7, trungtq2019, yalewkidane, kimd\}@kaist.ac.kr}
\and
\IEEEauthorblockN{Tae Joon Jun}
\IEEEauthorblockA{Asan Institute for Life Sciences\\
Asan Medical Center\\
Seoul, South Korea\\
taejoon@amc.seoul.kr}
}


%


\maketitle

\begin{abstract}
While advanced image captioning systems are increasingly describing images coherently and exactly, recent progress in continual learning allows deep learning models to avoid catastrophic forgetting. However, the domain where image captioning working with continual learning has not yet been explored. We define the task in which we consolidate continual learning and image captioning as continual image captioning. In this work, we propose \textbf{ContCap}, a framework generating captions over a series of new tasks coming, seamlessly integrating continual learning into image captioning besides addressing catastrophic forgetting. After proving forgetting in image captioning, we propose various techniques to overcome the forgetting dilemma by taking a simple fine-tuning schema as the baseline. We split MS-COCO 2014 dataset to perform experiments in class-incremental settings without revisiting dataset of previously provided tasks. Experiments show remarkable improvements in the performance on the old tasks while the figures for the new surprisingly surpass fine-tuning. Our framework also offers a scalable solution for continual image or video captioning.
\end{abstract}


%
\IEEEpeerreviewmaketitle

\section{Introduction and related work}
As a result of the success of deep learning over the last decade, image captioning has been rising as one of the most attractive domains because of its exceptionally valuable applications such as aiding to the visually impaired, social media, digital assistant, or photo indexing. Dealing with image captioning is to generate the most suitable caption describing the content of an image as much exact as possible. The most simple and prevalent architecture of an image captioning system is based on encoder-decoder architecture \cite{jia2015guiding, pu2016variational, donahue2015long}. In essence, the intermediate features are extracted by a Convolutional Neural Network (CNN) suppressed its output layer. The  features are then fed into an embedding block to synchronize with the input to a Recurrent Neural Network (RNN). The RNN, often powered by an LSTM, is in charge of generating a sentence word by word conditioned on the current time step and the previous hidden state. Sequence generation is performed until an end-of-sequence token is observed. Since the introduction of the Neural Image Caption Generator (NIC) \cite{vinyals2015show}, various powerful techniques have been proposed to enhance image captioning \cite{lu2018neural, cornia2019show, aneja2018convolutional}. Experiments are conducted on three benchmark datasets: Flickr8 \cite{hodosh2013framing}, Flickr30 \cite{plummer2015flickr30k}, and MS-COCO \cite{lin2014microsoft}. 

\begin{figure}[h!]
  \includegraphics[scale=0.32]{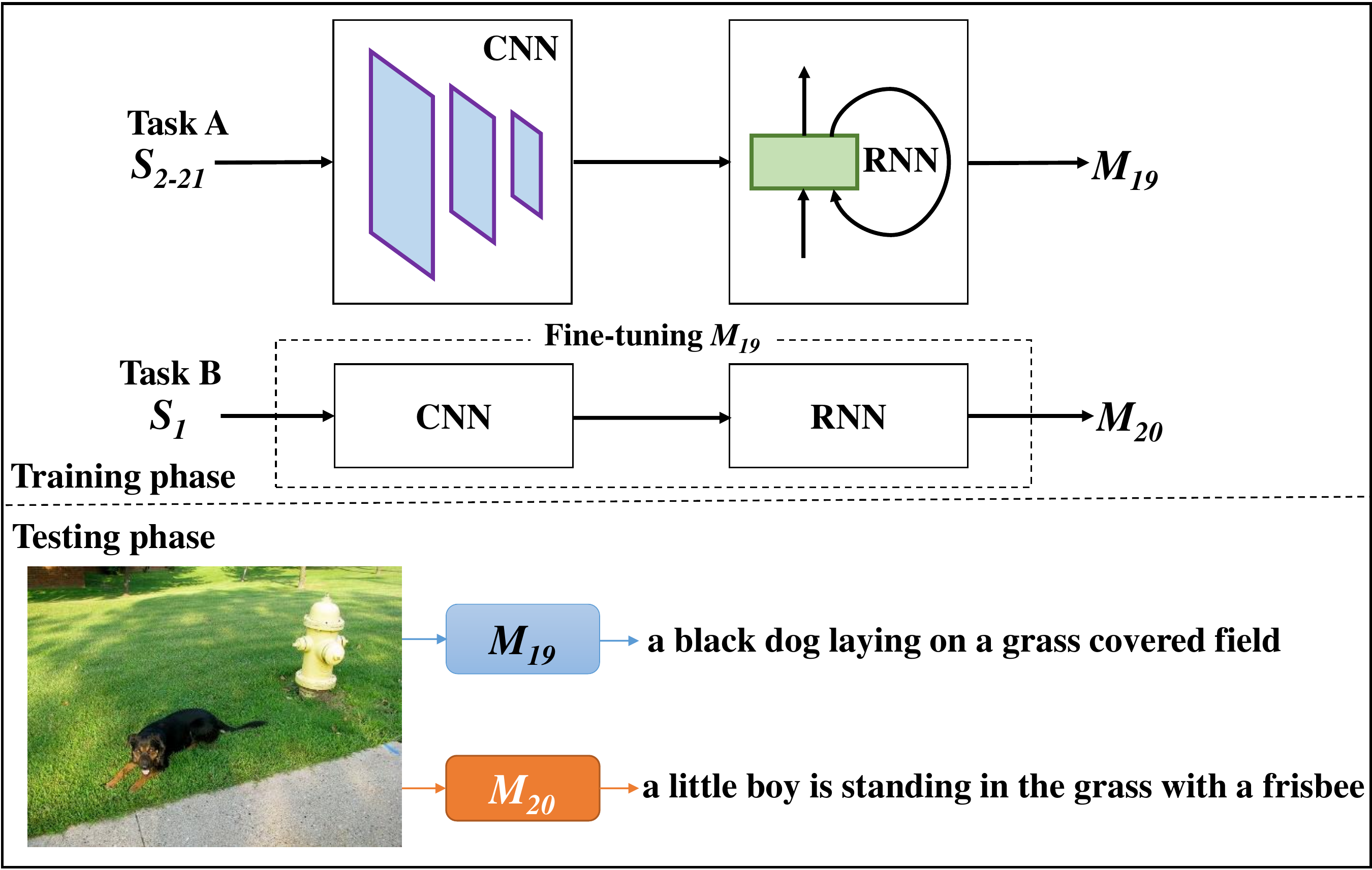}
  \caption{Catastrophic forgetting happens when adding a new class $person$.}
  \label{fig:forgetting}
\end{figure}

However, since image captioning models are built on the concept of Gradient-Based Neural Networks \cite{goodfellow2013empirical}, they inevitably suffer from catastrophic forgetting. A trained model can work well on a specific task thanks to the distribution it learned before. Nonetheless, it struggles to make use of this model on a new task while retaining its performance on the old task. More precisely, the model could reach a high performance on the new task, but its performance will be degraded disastrously on the old task. Though novel models \cite{xu2015show, lu2018neural} greatly improve the performance of image captioning, they have not considered catastrophic forgetting.

Continual learning requires an ability to learn over time without witnessing catastrophic forgetting (known as semantic drift), and it allows neural networks to incrementally solve new tasks. To achieve these goals of continual learning, studies mainly focus on addressing catastrophic forgetting problem. While continual learning community has not yet agreed on a shared categorization for continual learning strategies, Maltoni and Lomonaco \cite{maltoni2019continuous} propose a three-way fuzzy categorization of the most common continual learning strategies: Architecture, regularization, and rehearsal. Rehearsal approaches \cite{lopez2017gradient, tasar2018incremental} demand data from old tasks, leading to privacy or limited storage budget issues, making them non-scalable if tasks arrive ceaselessly. Architectural techniques \cite{rusu2016progressive, lopez2017gradient}, at the same time, alter the original network architecture for adapting to new tasks besides preserving old memory, but deteriorating the portability . The principle behind regularization approaches \cite{zenke2017continual, li2017learning} is to keep the model close to its previous state by adding a penalty to the objective function. Neither accessing old samples nor expanding the old network is required in regularization methods. Additionally, fine-tuning is a common technique used in deep learning, where the model can utilize what it has learned before to start solving a new problem. Ideas from continual learning are widely adopted for vision tasks like object detection and image classification \cite{shmelkov2017incremental, rebuffi2017icarl, castro2018end, wu2018incremental}. Very recently, Michieli and Zanuttigh \cite{michieli2019incremental} introduce a framework to perform continual learning for semantic segmentation. Our proposed techniques are based on the idea of regularization but slight expansions in the decoder are needed to maintain the preciseness of prediction.

Fig. \ref{fig:forgetting} shows catastrophic forgetting in image captioning when we adopt fine-tuning. Firstly, we train a model on a set of 19 classes $S_{2-21}$ (class 2: \textit{bicycle} to class 21: \textit{cow}) - called Task A to obtain the model $M_{19}$. The class 12 is missing in MS-COCO, so from class 2-21, we only have 19 classes. After that, we train on class 1: \textit{person} $S_{1}$ - Task B, fine-tuning from model $M_{19}$ to get the model $M_{20}$. Now to observe catastrophic forgetting, we pick an image from class 18: \textit{dog} in test set of Task A and drawn inference with both $M_{19}$ and $M_{20}$. With $M_{19}$, we call original model, the description is: \textit{a black dog laying on a grass covered field}. However, after fine-tuning to have $M_{20}$, the caption is shifted to: \textit{a little boy is standing in the grass with a frisbee}. Apparently, when learning the new task B, the model seriously forgets what it has learned before and seems to overemphasize describing \textit{person}. Even adding just one class, knowledge learned from old tasks is erased abruptly, causing a collapse in the performance. Continual learning is critical to image captioning especially when a real-time caption generator is being operated, ensuring captions are always strongly grounded on surrounding scenes.

In this paper, we propose \textbf{ContCap} that is a scalable framework to combine encoder-decoder image captioning architecture with continual learning. To overcome catastrophic forgetting, we introduce a pseudo-label approach which is an extended version from \cite{li2017learning} as our method works on the multi-modal architecture, while they work on CNN for image classification. The idea is to record the predictions of the antecedent model on the coming samples to tackle a new problem. Two new strategies of freezing (encoder or decoder) are also deployed to transfer the knowledge among tasks for preserving information of old tasks while adapting to work well on the new ones. Furthermore, we introduce distillation on the intermediate features (feature distillation) to guide the new model to produce a similar outcome to the old model. These techniques are applied separately for direct comparisons. The scalability of the framework is maintained as further continual learning techniques and more sophisticated encoder-decoder architectures can be smoothly integrated to increase the performance. Fine-tuning is considered as the baseline for comparison and evaluating the capability of strategies. 

Our paper addresses the most challenging scenario in continual learning where new training samples of different, previously unseen classes become available in subsequent tasks \cite{michieli2019incremental}. This scenario can be referred to as class-incremental learning \cite{lomonaco2019fine}. In particular, no access to data from old tasks is permitted, solving privacy concerns or memory obstacles. 

To conduct experiments in class-incremental settings, we create a new dataset from MS-COCO 2014 named Split MS-COCO in which each task will come with a new class. For example, with an old task A of classes {cat, dog, table}; the new tasks B will contain samples of class {person} only. When training task B, knowledge from task A is leveraged to help the current model generalize well on all 4 classes: {cat, dog, table, and person}. We define terminology ``clear image" to represent an image containing only one class out of 80 classes defined by MS-COCO in its reference captions to manipulate incremental steps correctly. 

The experiments demonstrate catastrophic forgetting in image captioning and the superiority of the proposed techniques over fine-tuning. The previously learned tasks are well captioned, whilst new information is also well absorbed. Traditional metrics in image captioning BLEU \cite{papineni2002bleu}, METEOR \cite{banerjee2005meteor}, ROUGE-L \cite{lin2004rouge}, and CIDEr \cite{vedantam2015cider} are calculated for quantitative assessments. Finally, we provide future directions and discussion over experiments to elaborate the results qualitatively and quantitatively.

\noindent
\textbf{Contributions:} Our main contributions are fourfold:
\begin{itemize}[noitemsep,nolistsep]
    \item Our work is the first attempt to address catastrophic forgetting in image captioning without the need for accessing data from existing tasks.
    \item We propose \textbf{ContCap} - a scalable framework that reconciles image captioning and continual learning in the class-incremental scenario. Our framework can be easily accommodated to further research in continual image captioning or video captioning, and effortlessly adopt towards existing advanced captioning models. 
    \item Our extensive experiments prove the manifestation of catastrophic forgetting in image captioning, and reveal particular effects of the freezing techniques, pseudo-labeling, and distillation in combating with this forgetting dilemma.
    \item We propose a new dataset named Split MS-COCO from the standard MS-COCO dataset. This can be a reference for other new benchmark datasets for continual learning. We particularize how to create Split MS-COCO in Section \ref{experiments}.
\end{itemize}

\section{Method} \label{Method}
\begin{figure*}[h!]
	\centering
	\includegraphics[scale=0.7]{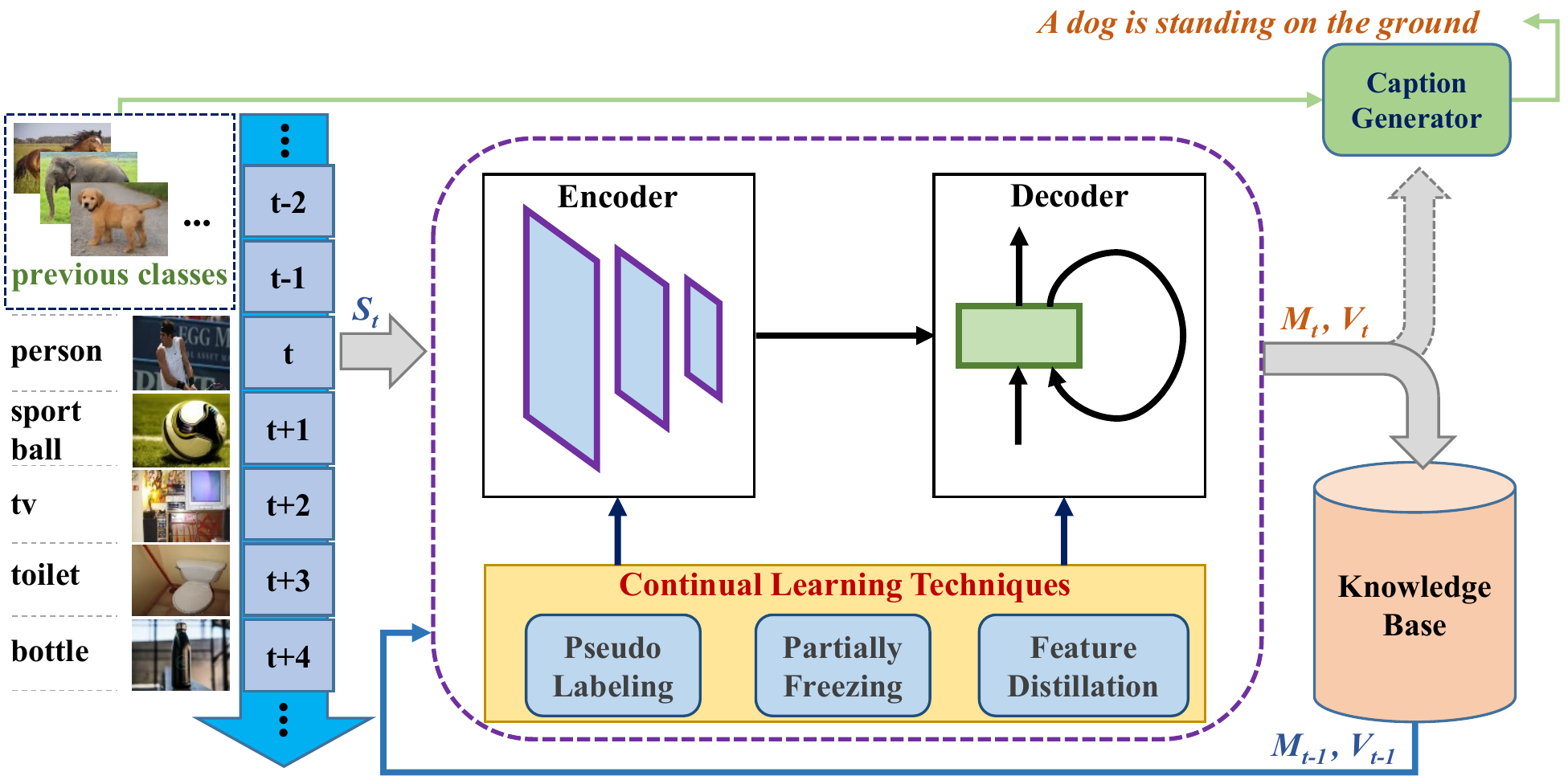}
	\caption{ContCap: A scalable framework for continual image captioning.}
	\label{fig:framework}
\end{figure*}

Given trained tasks followed by a series of new classes, the ultimate goal is to maximize the performance of captioning on the whole coming and previous tasks. The \textbf{ContCap} framework is presented in Fig. \ref{fig:framework}. At any time point $t$, the framework has performed a sequence of tasks ($T_{t-1}$,  $T_{t-2}$, ...), and samples from dataset of task $T_{t}$ are previously unseen. In learning task $T_{t}$, continual learning techniques are applied to information from Knowledge Base (KB) with an expectation to describe well on both the task $T_{t}$ and the previous ones. KB contains the acquired knowledge and the knowledge accumulated from learning the old tasks \cite{chen2016lifelong}. 

Derived from the network structure, we propose to freeze the encoder and decoder in turns (partially freezing) to moderate semantic drift. Pseudo-labeling makes use of the model $M_{k-1}$ to interpret predictions of the previous model on the coming samples while feature distillation utilizes $M_{k-1}$ as a teacher model, forcing student to not forget solved problems.

The proposed approaches can be applied over any encoder-decoder architecture-based image captioning models; however for evaluation, we pick the most popular architecture purely constructed by CNN + LSTM. The encoder for feature extraction is the pre-trained ResNet-152 \cite{he2016deep}. For the decoder part, we build a single layer LSTM network for managing the process of description generation. A task $T$ is denoted by its classes $C$, data $S$, obtained model $M$ and vocabulary $V$. Initially, we train the network on a set of classes ($C_{t-1}$, $C_{t-2}$, ...) with the corresponding data ($S_{t-1}$, $S_{t-2}$, ...) to get the model $M_{t-1}$. Next, with incremental steps $k$ = $t$, $t+1$, $t+2$, ..., in task $T_{k}$, we load the model $M_{k-1}$ and update the current model employing the distribution of $S_{k}$.

Although tasks coming later are independent from previous tasks, a vocabulary should be accumulated and transferred progressively. In fact, in an early task of 19 classes, we have a vocab $V_{19}$ of $v_{19}$ entries. When a task of \textit{person} comes, we simply perform tokenization by NLTK \cite{loper2002nltk} on the annotations of samples, then pick the most significant words into the vocabulary. The accumulated vocabulary so far is $V_{20} = V_{19} \cup V_{person}$, and the number of entries in the vocabulary is $v_{20} = v_{19} + v_{person} - v_{V_{19} \cap V_{person}}$.

We run the fine-tuning experiment at the first step to get the baseline for comparison. The remaining techniques are then separately applied. With new classes, we initialize the model parameters by the converged state of the last task. Predictably, the new model will try to generalize from the newly added classes and give a poor performance on the old task by default. Models with continual learning approaches, with the exception of feature distillation, are initialized by the weights of the previous task prior to training.

\subsection{Partially freezing for image captioning}
A heuristic solution to hinder catastrophic forgetting is to freeze a part of model. As our architecture is divided into an encoder and a decoder, keeping one component intact can affect directly to the performance on the previous tasks as well as the new task. 

\begin{figure}[h!]
  \includegraphics[scale=0.330]{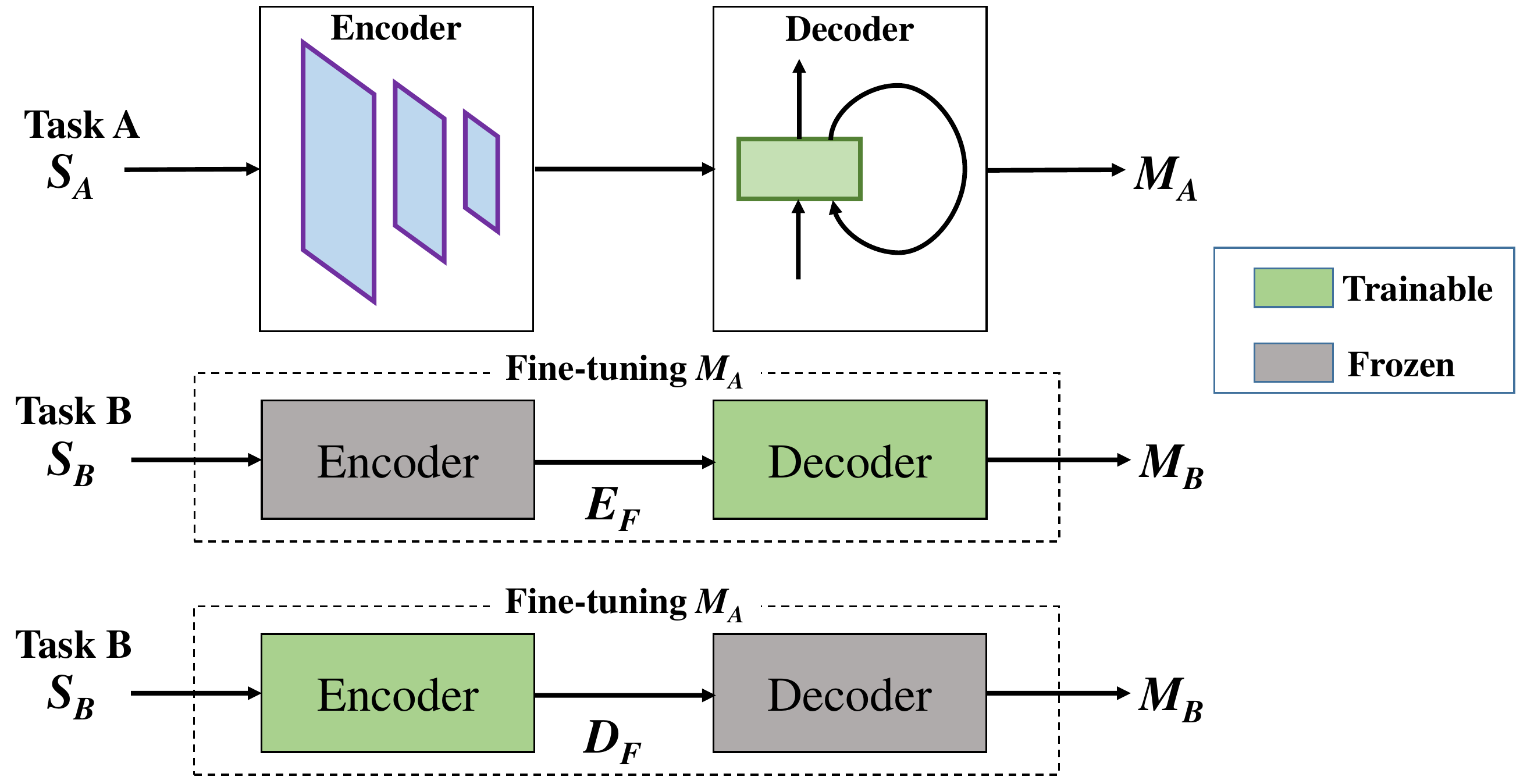}
  \caption{Freezing a component of the fine-tuned model while the remaining component is trainable.}
  \label{fig:freezing}
\end{figure}

On the one hand, the encoder is frozen and the decoder is updated exploiting the new dataset. The feature extraction remains compared to the previous model $M_{A}$ which operates as a fine-tuned model (Task A). The feature extractor $E$ is constrained, leaving combatting with the new task for the decoder $D$, we refer as $E_{F}$ (see Fig. \ref{fig:freezing}). The vocabulary is expanded to ensure the naturalness of the prediction, and the decoder is modified according to the vocabulary size. 

On the other hand, we freeze the decoder $D$ while the encoder is trainable. Because neurons are added to the decoder as by virtue of expanding vocabulary, the newly attached neurons are the only trainable part of the decoder. By setting like this, the convolutional network can better learn unseen features, thus feeding more fine-grained input into the decoder. In the case of both $E$ and $D$ being frozen, the new model is the same as the previous model, while fine-tuning is when both $E$ and $D$ are trainable.

The objective function when applying freezing techniques is the standard loss function for image captioning:

\begin{equation}
\label{eq:1}
\begin{aligned}
	\mathcal{L} ={} & L_{CE}\\
	            ={} & - \sum_{i=1}^{v} Y_{k}^{i} \log \hat{Y}_{k}^i
\end{aligned}
\end{equation}
In the equation \eqref{eq:1}, $L_{CE}$ is the cross-entropy loss over annotation and the prediction, $v$ is the size of vocabulary. $Y_{k}$ is the ground truth (one-hot vector), and $\hat{Y}_{k}$ is the predicted caption (probability distribution over vocabulary). 

\subsection{Pseudo-labeling}
We refer this method as pseudo-labeling since when training, pseudo-labels are generated to guide the current model to mimic the behavior of the previously trained model. The procedure of this approach is described as follows:

\begin{lstlisting}[mathescape,escapechar=\%]
At task $k^{th}$, 
%\underline{\textit{Start with}}%:
  $\theta_{k-1}$ %\space\space\space: parameters of the old task%
  $X_k$, $Y_k$ %: training images and reference captions on the new task%
%\underline{\textit{Initialize}}%:
  $Y_{k-1}$ $\leftarrow$ $M(X_{k},\theta_{k-1})$ %\space// infer captions by the old%
                        // model on new data
  $\theta_k \leftarrow \theta_{k-1}$ %\space\space\space\space\space\space\space\space\space\space\space\space\space\space// initialize new model by old%
                        // parameters
%\underline{\textit{Train}}%:
  $\hat{Y}_k \leftarrow M(X_k,\hat{\theta}_k)$ %\space\space\space\space\space\space// new task prediction%
  $\theta_k^{*} \leftarrow \underset{\hat{\theta}_k}{argmin}(L(Y_k,\hat{Y}_k) + L(Y_{k-1},\hat{Y}_k))$
\end{lstlisting}

In \cite{li2017learning}, they use task-specific parameters to generate the pseudo labels while in our proposed pseudo-labeling approach, we use a unified network for the whole process. This point eases the training because we do not need to separate the network into sub-modules. We also do not apply knowledge distillation in pseudo-labeling approach because we do not expect the logit values in captioning task. The pseudo labels are fake captions on the new data.

Pseudo labels $Y_{k-1}$ are acquired by inferring the caption of all input images in the dataset of the new task using the previous model. Although new classes appear, the convolutional network architecture stays unchanged during training since the expected output is textual caption, not the probabilities for classes as in object classification or semantic segmentation. After being initialized by the foregoing model, we run training in a supervised manner to minimize the loss computed based on the ground truth, the predicted caption, and the pseudo labels $Y_{k-1}$. The loss component from pseudo-labeling $L_{P}$ is explicitly written as:

\begin{equation}
\begin{aligned}
    {L_{P}} = - \beta \cdot \sum_{i=1}^{v} Y_{k-1}^i \log \hat{Y}_{k}^i
\end{aligned}
\end{equation}
where $\beta$ is considered to be a regulator to accentuate the old or new task ($\beta = 1$ in experiments), and the final objective function for this approach is:
\begin{equation}
\begin{aligned}
	\mathcal{L} ={} & L_{CE} + L_{P}\\
	            ={} & - \sum_{i=1}^{v} Y_{k}^{i} \log \hat{Y}_{k}^i - \beta \cdot \sum_{i=1}^{v}Y_{k-1}^i \log \hat{Y}_{k}^i
\end{aligned}
\end{equation}

\subsection{Feature distillation}
This technique frames the training process in a teacher-student strategy when we want to deliver the knowledge acquired from one model to another model. By applying distillation, teacher is the model obtained from the old task while student is the new model. Once an image $X_{n}$ is passed through both teacher and student, the mean squared error between outputs of student and teacher ($Y_{k}^{st}$ and $Y_{k}^{tr}$) is added to the loss function of student to penalize it. We propose to distill the feature learned by the encoder to provide the student decoder the flexibility adapting to the new task. This approach is felicitous because the spatial information is global while the sentence semantics changes constantly over domains. Thus, the decoder is freer to change, in comparison with the encoder. The distillation term $L_{dis}$ is computed as:
\begin{equation}
\begin{aligned}
	{L_{dis}} = \lambda\cdot\left\| Y_{k}^{tr} - Y_{k}^{st} \right\|^2 
\end{aligned}
\end{equation}
where $\lambda$ may be increased to encourage student to imitate the behaviors of teacher more intensively ($\lambda = 1$ in experiments). The loss function is:
\begin{equation}
\begin{aligned}
	\mathcal{L} ={} & L_{CE} + L_{dis}\\
	            ={} & - \sum_{i=1}^{v} Y_{k}^{i} \log \hat{Y}_{k}^{i}  + \lambda\cdot\left\| Y_{k}^{tr} - Y_{k}^{st} \right\|^2 
\end{aligned}
\end{equation}

\section{Experimental result}\label{experiments}
\paragraph{Dataset} Our experimental results have been recorded on Split MS-COCO, a variant of the MS-COCO benchmark dataset \cite{lin2014microsoft}. The original MS COCO dataset contains 82,783 images in the training split and 40,504 images in the validation split. Each image of MS COCO dataset belongs to at least one of 80 classes. To create Split MS-COCO, we carry out processing on MS-COCO as the following steps:

\newenvironment{packed_enum}{
\begin{enumerate}
  \setlength{\itemsep}{1pt}
  \setlength{\parskip}{0pt}
  \setlength{\parsep}{0pt}
}{\end{enumerate}}

\begin{packed_enum}
    \item Split all images into distinguished classes. There are 80 classes in total.
    \item Resize images to a fixed size of ($224\times224$) in order to deal with different sizes of input.
    \item Discard images including more than one class in its annotations to obtain ``clear" images only.
\end{packed_enum}

\begin{table*}[ht!]
\centering
\caption{Performance of $M_{20}$ on $S_{19}$ and $M_{20}$ on the new class $person$.}
\begin{adjustbox}{width=1.0\textwidth,center=\textwidth}
\begin{tabular}{|l|c|c|c|c|c|c|c|c|c|c|}
\hline
\multirow{2}{*}{} & \multicolumn{5}{c|}{$S_{19}$}                         & \multicolumn{5}{c|}{$person$}                                                  \\ \cline{2-11} 
                  & BLEU1          & BLEU4          & METEOR         & ROUGE\_L       & CIDEr          & BLEU1          & BLEU4          & METEOR         & ROUGE\_L       & CIDEr          \\ \hline
$F$       & 47.3         & 5.8        & 11.3        & 35.2         & 9.4         & 62.6         & 19.2        & 20.0 & 46.6         & 43.6         \\ \hline
$E_{F}$    & 48.2         & 7.5         & 12.3          & 35.4          & 14.3          & 60.7 & 18.4         & 19.6          & 45.4         & 40.4 \\ \hline
$D_{F}$    & 47.9        & 6.9        & 12.4        & 35.3          & 16.3          & 57.1        & 14.4 & 17.5         & 43.4 & 26.3        \\ \hline
$P$     & \textbf{52.7} & \textbf{10.7} & \textbf{14.5} & \textbf{39.0} & \textbf{25.3} & 58.8          & 16.4         & 18.2         & 44.5         & 32.6          \\ \hline
$FD$     & 50.7 & 8.0 & 12.7 & 36.6 & 16.2 & \textbf{65.0}          & \textbf{22.3}         & \textbf{21.5}         & \textbf{48.1}          & \textbf{52.7}          \\ \hline
\end{tabular}
\end{adjustbox}
\label{tab:my-table1}
\end{table*}

\begin{table*}[ht!]
\centering
\caption{Performance of $M_{24}$ on $S_{19}$ and $M_{24}$ on $S_{multiple}$ (classes $person$, $sports$ $ball$, $tv$, $toilet$, and $bottle$ are added at once).}
\begin{adjustbox}{width=1.0\textwidth,center=\textwidth}
\begin{tabular}{|l|c|c|c|c|c|c|c|c|c|c|}
\hline
\multirow{2}{*}{} & \multicolumn{5}{c|}{$S_{19}$}                         & \multicolumn{5}{c|}{$S_{multiple}$}              \\ \cline{2-11} 
                  & BLEU1          & BLEU4          & METEOR         & ROUGE\_L       & CIDEr          & BLEU1          & BLEU4          & METEOR         & ROUGE\_L       & CIDEr          \\ \hline
$F$       & 45.4         & 6.9        & 11.3          & 34.4         & 9.8          & 63.3         & 21.6         & 20.7          & 47.6        & 54.2         \\ \hline
$E_{F}$    & 49.1         & 8.7         & 12.3         & 36.4         & 15.6         & 60.1         & 18.7         & 18.3 & 45.0         & 43.4 \\ \hline
$D_{F}$     & 49.8         & 9.1         & 13.1         & 36.7          & 20.8         & 59.4 & 17.9 & 18.4          & 45.0 & 38.0          \\ \hline
$P$     & 51.2 & \textbf{11.0} & \textbf{14.3} & 38.1 & \textbf{25.1} & 59.2        & 17.8          & 18.1         & 45.0         & 38.2         \\ \hline
$FD$     & \textbf{51.7} & 10.7 & 14.0 & \textbf{38.3} & 22.8 & \textbf{65.7}        & \textbf{23.5}         & \textbf{22.0}          & \textbf{49.3}       & \textbf{60.3}         \\ \hline
\end{tabular}
\end{adjustbox}
\label{tab:my-table2}
\end{table*}

\begin{table*}[ht!]
\centering
\caption{Performance of $M_{seq}$ on $S_{19}$ and $M_{seq}$ on $S_{multiple}$ (classes $person$, $sports$ $ball$, $tv$, $toilet$, and $bottle$ are added at one by one).}
\begin{adjustbox}{width=1.0\textwidth,center=\textwidth}
\begin{tabular}{|l|c|c|c|c|c|c|c|c|c|c|}
\hline
\multirow{2}{*}{} & \multicolumn{5}{c|}{$S_{19}$}                         & \multicolumn{5}{c|}{$S_{multiple}$}              \\ \cline{2-11} 
                  & BLEU1          & BLEU4          & METEOR         & ROUGE\_L       & CIDEr          & BLEU1          & BLEU4          & METEOR         & ROUGE\_L       & CIDEr          \\ \hline
$F$       & 42.7          & 4.0          & 9.2          & 32.2          & 4.5          & 49.8          & 9.0          & 12.8          & 39.0         & 13.2          \\ \hline
$E_{F}$    & 43.8         & 5.2          & 10.9        & 33.0          & 7.8          & 50.8          & 10.3          & \textbf{14.5}          & 39.3          & \textbf{19.2}          \\ \hline
$D_{F}$    & 45.6 & 5.0 & \textbf{11.2} & 33.7 & 9.9 & 51.3 & 9.5 & 13.6 & 38.9 & 12.3 \\ \hline
$P$     & 46.3          & \textbf{6.6}         & 10.6         & 32.6         & \textbf{10.0}          & 53.4          & \textbf{10.5}          & 13.9          & 39.5          & 15.1          \\ \hline
$FD$     & \textbf{47.1} & 5.1 & 10.7 & \textbf{34.0} & 4.8 & \textbf{53.6}            & 9.4        & 14.3           & \textbf{40.0}        & 13.3          \\ \hline
\end{tabular}
\end{adjustbox}
\label{tab:my-table3}
\end{table*}

Taking a clear ``dog" image as an example, the reference captions will contain ``dog" in their content and may also have ``grass" (not in 80 classes) but NOT contain any of the 79 remaining classes defined by MS-COCO (cat or frisbee). The reason that we only choose clear images is that we want to assure when facing a new task, objects in images of the new task are unseen. 

Since MS-COCO test set is not available, we divide the validation set of each class into two equal parts: one is for validation and the other for testing. From over 82k images of training set and 40k of validation set from the original MS-COCO 2014, the new dataset has 47,547 images in the training set; 11,722 images in the validation set; and 11,687 images in the test set.

\paragraph{Evaluation metric} We report the capacity of the framework using traditional scores for image captioning, such as BLEU, ROUGE, CIDEr, and METEOR by \texttt{coco-caption} \cite{chen2015microsoft}. $F$, $E_{F}$, $D_{F}$, $P$, and $FD$ are fine-tuning, freezing encoder, freezing decoder, pseudo-labeling, and feature distillation, respectively. While BLEU, ROUGE, and METEOR are variants of overlapping-based metrics and linearly correlated, CIDEr gives more weight-age to important terms, thus giving higher correlation with human judgments. From that, we will only mention BLEU and CIDEr in the following for short. For each scenario, we have a table correspondingly. The performance of the techniques on the old task is on the left side while the figure for the new task is located on the right side of the tables.

\subsection{Addition of one class}
In the first setting, we examine catastrophic forgetting by an addition of one class. We train the model $M_{19}$ with the training set of 19 classes (2 to 21, but class 12 is not available, so we only have 19 classes) - $S_{19}$, and then evaluate on the corresponding test set of these 19 classes. Next, we analyze the generalization capacity of the model by adding another class $person$ and see how confident the model $M_{20}$ can execute on the old and new tasks. The results are shown in the Table \ref{tab:my-table1}.

From the table, we can argue that fine-tuning seriously demolishes the accuracy of the old task although we just add only one class. The old model $M_{19}$ is almost broken. Massive decreases can be observed in all the metrics with the most remarkable one observed in CIDEr, a drop from 47.3 to 9.4. On the old task, pseudo-labeling outperforms the remaining ones, reaching 25.3 CIDEr, which nearly triples as fine-tuning's CIDer score. Freezing encoder and decoder achieve 14.3 and 16.3 CIDEr respectively. We conclude that freezing does not help much because when a part is frozen, the other might be strongly driven to the new domain, thus misdescribing images of the old domain. Pseudo-labeling improves retaining the old knowledge as the pseudo-labels impede the drift of model weights by the regularization. Feature distillation performs moderately because we rebuild the model, not any information is given from the beginning, standing at only 16.2 CIDEr.

When testing on the new domain $person$, feature distillation reaches a relatively high performance (65.0 and 52.7 for BLEU1 and CIDEr respectively), surprisingly surpassing fine-tuning (62.6 BLEU1 and 43.6 CIDEr). This is due to the fact that the global features inherited from the teacher compromises with the new features, resulting in richer representation. Simply freezing decoder performs worst while pseudo-labeling yields 58.8 BLEU1 and 32.6 CIDEr. Despite an average performance on the old task of feature distillation, this technique might be the most reasonable option when adding a new class as it successes on the new task and still improves the old task performance. Another finding is that freezing encoder is more powerful than keeping decoder frozen.

\subsection{Addition of multiple classes at once}
We assume that the dataset formed by these 5 classes $person$, $sports$ $ball$, $tv$, $toilet$, and $bottle$ is $S_{multiple}$. The implication for picking those 5 classes is that they are the biggest classes that are not present in $S_{19}$, and they are from different super-classes \cite{lin2014microsoft} (ranging from animal to kitchen), ensuring totally new instances are introduced.
In this setting, starting with $M_{19}$ and $S_{19}$, the new task is represented by $S_{multiple}$. After training to get the model $M_{24}$, we test the model on both $S_{19}$ and $S_{multiple}$. The scores are reported in the Table \ref{tab:my-table2}.

We first test $M_{24}$ over $S_{19}$. As the vocabulary size has been broadened significantly due to new words coming from the annotations of the whole 5 new classes, the generated sentences take a chance to be more natural and diverse. As a result, all approaches in this experiment perform better than the ones when only one class is added. 
On both the old and new task, the trend of results is roughly similar to the first experiment when we add a new class $person$. We suppose that since the 5 new classes belong to different and non-existing super classes, the features from each of them are unique, thus generally similar to adding a new class. We refer this type of adding as ``unknown-domain addition". A domain here means a super class such as $person$, $sports$, $furniture$, or $animal$. Noticeably, only pseudo-labeling aggravates the performance on  $S_{19}$. We argue that although $M_{19}$ forces $M_{24}$ to generate identical outputs, $M_{19}$ utilizes $V_{19}$ to generate fake captions. Therefore, that using the new vocabulary to generate words in the old vocabulary makes noise in the training process. Adding more classes at once while using pseudo-labeling may not bring a better generalization.

Adding multiple classes simultaneously is similar to adding a new class but generates more fine-grained descriptions thanks to a bigger vocabulary. 
While pseudo-labeling has the performance balance (25.1 on $S_{19}$ and 38.2 CIDEr on $S_{multiple}$), feature distillation is still the leading technique in this scenario, at 22.8 and 60.3 CIDEr. Keeping encoder fixed is once again proven to have an equal effect to doing on decoder (15.6 CIDEr and 43.4 CIDEr compared to 20.8 and 38.0) in this scenario.

\subsection{Addition of multiple classes sequentially}
Again, we begin with $M_{19}$ and $S_{19}$. After that, we add 5 classes $person$, $sports$ $ball$, $tv$, $toilet$, and $bottle$ one by one. At each stage, the previous model is used to initialize the current model. The class of $bottle$ is trained at the end of the process. Finally, we have $M_{seq}$ to carry out evaluation on $S_{19}$ and $S_{multiple}$. 

By looking at the Table \ref{tab:my-table3}, we realize that this setting causes extremely poor performance on $S_{19}$ regardless of the presence of the techniques.
BLEU1 scores which are computed in 1-gram manner stand at around 45.0, but other metrics collapse because they count matching n-grams or even more complex patterns. Sequentially adding new classes drifts the original model to various directions after each step, thus changing it drastically. We consider this problem as ``stage catastrophic forgetting" which is an instance of catastrophic forgetting but especially appears when unseen classes arrive incrementally. At the end of the addition, everything learned from the old task is completely erased. CIDEr decreases from 47.1 to only 4.8 with feature distillation and 42.7 to 4.5 with fine-tuning. Besides, pseudo-labeling and freezing decoder show the same performance, standing at around 10.0 CIDEr.

Regarding $S_{multiple}$, the performance is much lower than the figure for the setting of adding 5 classes at once. Simultaneously adding helps models better generalize the knowledge on new classes, while sequentially adding suffers from both stage catastrophic forgetting and poor generalization. Interestingly, freezing encoder beats freezing decoder (7.8 CIDEr and 19.2 CIDEr compared to 9.9 and 12.3), again suggesting that encoder has a more significant role than decoder in learning a long range of tasks.
Although being exceeded by freezing encoder, pseudo-labeling (53.4 BLEU1 and 15.1 CIDEr) still works well and outperforms both fine-tuning and feature distillation.
We believe that pseudo-labeling dominates this scenario because stage catastrophic forgetting overshadows the noise from the accumulated vocabulary. Hence, it is more durable over the range of tasks, whereas other techniques fluctuate and show a poor generalization.

\subsection{Qualitative analysis} 
\begin{figure*}[h!]
	\centering
	\includegraphics[scale=0.65]{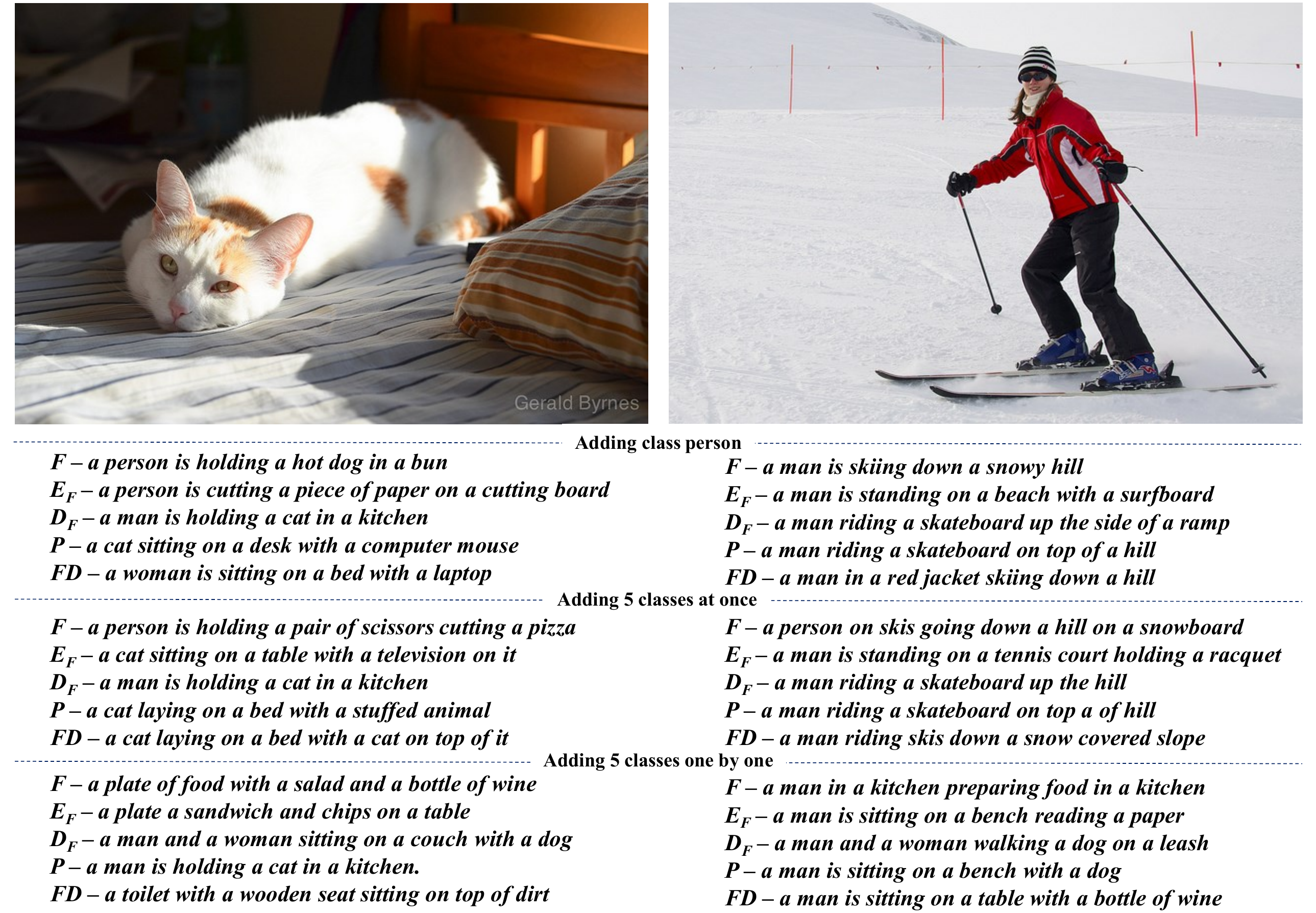}
	\caption{Qualitative results on samples from the old and new tasks. Three scenarios are presented here illustrate that the qualitative results are correlated with the computed scores. The impacts of adding one class and multiple classes at once are approximately similar, while the quality of caption in sequential addition declines catastrophically after each step.}
	\label{fig:qualitative_result}
\end{figure*}

Captions are inferred in every setting to qualitatively evaluate the framework (See Fig. \ref{fig:qualitative_result}). The model misclassifies a cat without using pseudo-labeling when adding a class $person$. Adding 5 classes at once helps the model accomplish a better generalization, so descriptions are more aligned to the content of images. Predictions become ambiguous when we have 5 classes sequentially in all the old and new tasks because the network is driven to many directions, causing a difficulty in reaching an optimal point for the model. 

\section{Conclusion}
In this paper, we introduce \textbf{ContCap}, a scalable framework fusing continual learning and image captioning, working on a new dataset called Split MS-COCO created from the standard MS-COCO. We firstly perform image captioning in incremental schemes and then subsequently propose algorithms to attenuate catastrophic forgetting. Working on the most challenging scenario (class incremental) asserts our framework can perform well in practical schemes while old classes and new classes can co-occur. The experiments in three settings indicate the sign of catastrophic forgetting and the effectiveness when integrating freezing, pseudo-labeling, and feature distillation compared to fine-tuning.  

Applying further advanced techniques from continual learning to enhance the generality especially in the sequential addition setting is left for future work. Since the image captioning task is multi-modal, further approaches should be devised to reach a high performance like single-modal tasks (object classification or segmentation). As we mentioned ``unknown-domain addition", experiments with ``known-domain addition" should be conducted in an expectation of witnessing less forgetting compared to ``unknown" scenario. Stage catastrophic forgetting should be more deeply investigated to make the learning process possible when we have a stream of new tasks. 

\bibliography{example_paper}
\bibliographystyle{IEEEtran}

\end{document}